\documentclass[11pt]{article}
\usepackage[final]{acl}
\usepackage{times}
\usepackage{latexsym}
\usepackage[T1]{fontenc}
\usepackage[utf8]{inputenc}
\usepackage{microtype}
\usepackage{inconsolata}
\usepackage{graphicx}
\usepackage{amsmath}
\usepackage{booktabs}
\usepackage{placeins}

\newcommand{\PaperRepositoryAvailability}{%
  The code and supplementary materials are available in a public GitHub
  repository.\footnote{\url{https://github.com/DKT-PROJECTS-UCLA/KC_Clusters}}}
\newcommand{\PaperRepositoryReference}{the public GitHub repository}

\title{Embedding-Based Educational Taxonomies}
\author{%
  \textbf{Kia Karbasi\textsuperscript{1}},
  \textbf{Kaveh Karbasi\textsuperscript{2}\thanks{Tempus AI, Inc. had no role in the
  planning, execution, funding, or review of this work.}},
  \textbf{Gregory Pottie\textsuperscript{1}} \\
  \normalfont
  \textsuperscript{1}University of California, Los Angeles
  \quad
  \textsuperscript{2}Tempus AI, Inc.}

\begin{document}
\maketitle

\begin{abstract}
Educational taxonomies organize educational content, such as exercise problems,
into concepts, which are also called skills or knowledge components. These
taxonomies define learning objectives in educational applications such as
intelligent tutoring systems. Developed through human expertise and iterative
refinement, they also provide fundamental structure for curricula, assessment,
and models of student learning. Recent advances in semantic text embeddings
make it possible to construct alternative educational taxonomies computationally
from the content itself. In this paper, we investigate this possibility by
clustering embeddings of math questions to construct
embedding-based educational taxonomies. We compare these taxonomies with the
existing Common Core State Standards (CCSS)-based knowledge-component taxonomies
in two ASSISTments datasets
and evaluate their utility in Deep Knowledge Tracing, a downstream task that
relies on taxonomic mappings of questions to knowledge components. The
embedding-based taxonomies exhibit substantial but incomplete correspondence
with the existing taxonomies, revealing both preserved and alternative
organizations of educational content. When used for knowledge tracing, they
improve predictive performance by approximately 3--5 AUC percentage points.
These results demonstrate the potential of semantic embeddings for automatically
constructing educational taxonomies that both recover structure present in
human-developed taxonomies and provide useful alternative representations for
downstream educational modeling.
\end{abstract}

\section{Introduction}

Educational taxonomies organize educational content, such as exercise problems,
into concepts. In educational data and modeling, these concepts are often called
skills or knowledge components (KCs). A taxonomy specifies which questions are
treated as instances of related knowledge, giving educators and computer
systems a common way to describe content and learning objectives.
These mappings provide structure for curricula, assessment, educational
applications, and models of student learning~\cite{
koedingerKnowledgeLearningInstruction2012}.

Educational taxonomies are generally developed through human expertise and
iterative refinement. The Common Core State Standards, for example, were
developed through research, educator and expert input, comparison with existing
standards, and public feedback~\cite{ccssDevelopmentProcess2010}. Such systems
encode decisions about which distinctions matter for curricula and assessment,
while KC models use those distinctions to relate educational tasks, instruction,
and student learning~\cite{porterCommonCoreStandards2011,
koedingerKnowledgeLearningInstruction2012}. They are therefore useful educational
frameworks, not direct evidence that knowledge is organized in one uniquely
correct way.

These taxonomies also provide the categories that educational technologies use
to organize content and model student learning. Intelligent tutoring systems
connect educational materials to the concepts they are intended to teach or
assess~\cite{
koedingerKnowledgeLearningInstruction2012,fengAddressingAssessmentChallenge2009}.
Knowledge tracing uses sequences of questions and responses, along with the
concepts assigned to those questions, to estimate student knowledge and predict
later performance~\cite{corbett_knowledge_1995,
abdelrahmanKnowledgeTracingSurvey2023}. Deep Knowledge Tracing (DKT) represents
each response with its assigned KC, so earlier performance on a KC can inform
later predictions~\cite{piech_deep_2015}. Systems that recommend exercises
similarly use concept labels and estimates of student knowledge to select
individualized practice~\cite{
aiConceptAwareRecommendation2019}.

Recent advances in text embeddings make it possible to organize educational
content using patterns in the content itself. Embeddings represent question text
in a numerical space where similarities in language and meaning can be
compared~\cite{reimers_sentence-bert_2019}. Clustering these representations
groups nearby questions and assigns each question to one category~\cite{
matsudaLatentSkillMining2022}.

In this paper, we use semantic embeddings of math
question text to induce embedding-based taxonomies. Here, a taxonomy refers to a
mapping between content or questions and content categories or knowledge
components. An induced taxonomy is one constructed from text embeddings. We
compare these induced taxonomies with the original
expert-developed taxonomies and test their utility in a downstream educational
task that relies on question-to-KC mappings.

We study questions with available text from ASSISTments 2012 and 2017, two
releases from a widely used benchmark family in knowledge tracing~\cite{
fengAddressingAssessmentChallenge2009,priharExploringCommonTrends2022,
abdelrahmanKnowledgeTracingSurvey2023}. We embed the question text, reduce the
representations with principal component analysis (PCA), and cluster them with
\(k\)-means to construct taxonomies from embeddings. We first compare each
taxonomy built from embeddings with the original KC taxonomy supplied with the
dataset, using several measures of how the questions are grouped, along with
category counts. We then replace the original mapping from questions to KCs with
the induced mapping and compare DKT performance using the area under the receiver
operating characteristic curve (AUC). Within each dataset, we use the same
student-response records, sequence order, student split, DKT model family,
training procedure, Bayesian hyperparameter search, and trial budget. Because an induced
taxonomy may use a different number of categories, the number of KC labels and
the number of model parameters tied to those labels may also differ. We tune DKT
separately for each taxonomy, so the selected DKT hyperparameters may differ.
\PaperRepositoryAvailability

We ask three research questions:

\textbf{RQ1:} How closely do the induced and original KC taxonomies correspond?

\textbf{RQ2:} How does correspondence change as the induced taxonomy uses more
categories?

\textbf{RQ3:} Under which taxonomy does DKT achieve higher AUC: the original KC
taxonomy or the embedding-induced taxonomy?

\section{Related Work}

\subsection{Educational Taxonomies and Knowledge Components}

KC models connect performance on related tasks to proposed units of knowledge
used in teaching and models of student learning~\cite{
koedingerKnowledgeLearningInstruction2012}. A mapping from questions to KCs
therefore represents a hypothesis about the knowledge required by each question.
These models are designed for educational use; they are not direct measurements
of how students think. Learning Factors Analysis shows that student data can be
used to evaluate and refine an expert KC model~\cite{
cenLearningFactorsAnalysis2006}. Whereas that approach compares possible
refinements using student performance, we construct alternative categories from
question content. We therefore use the existing KC taxonomy as a valuable
reference, not as the only possible way to group the questions. Expert-developed
taxonomies reflect instructional organization; induced taxonomies offer complementary
groupings derived from question text. Whether these groupings align with how
students organize or learn that content remains an open question.

\subsection{Embedding-Based Clustering and Automated KC Discovery}

Sentence embeddings represent text in a space that supports semantic comparison
and clustering~\cite{reimers_sentence-bert_2019}. Text similarity can
suggest shared content, but it does not by itself establish a common solution
method or role in teaching. SMART uses \(k\)-means to cluster representations of
instructional and assessment text in which terms are weighted by their
frequency within a document and rarity across documents (TF--IDF), compares the
induced categories with expert categories, and evaluates the resulting mapping with an
Additive Factors Model (AFM)~\cite{matsudaLatentSkillMining2022}. KCluster uses
language-model-based question similarity and affinity propagation to derive KCs,
compares them with an expert taxonomy, and also evaluates them with AFM~\cite{
weiKClusterLLMBased2025}. These are the clustering studies most closely related
to our work, although neither uses ASSISTments or DKT. Moon et al.\ extract KCs
from question text and images, group them using embeddings, and evaluate the
resulting categories with several knowledge-tracing models, including
DKT~\cite{moonLargeMultimodalKC2025}.

Together, these studies show that educational categories can be constructed from
course and question content. Our study builds on this work by comparing induced
and original taxonomies across every tested combination of PCA dimension and number of induced
categories (\(K\)), then evaluating the induced mappings in DKT. We also ask
whether settings where the taxonomies match more closely tend to produce higher
AUC. The core methodology was developed independently of SMART and KCluster,
which we identified as closely related approaches during subsequent literature
review.

\subsection{Knowledge Tracing and Text-Aware Student Modeling}

Knowledge tracing uses question--response histories to model how student
knowledge develops over time~\cite{corbett_knowledge_1995,
abdelrahmanKnowledgeTracingSurvey2023}. DKT uses a recurrent neural network to
predict later responses from earlier responses and their KC labels~\cite{
piech_deep_2015}. For example, if two questions share a KC label, DKT can use
a student's earlier performance on that KC when predicting a later response;
under our method, the same process uses the induced KC labels instead. Methods
that use question text add question representations to
student models while retaining the existing skill labels~\cite{
su_exercise-enhanced_2018,liu_ekt_2019}. Our approach uses embeddings before DKT
is trained: the embeddings define an alternative mapping from questions to
categories, and DKT is then trained with the induced labels. This lets us
evaluate the mapping in two ways: how closely it matches the original taxonomy
and the test AUC achieved by DKT with the induced labels.

\section{Method}

\begin{figure*}[t]
\centering
\includegraphics[width=.98\textwidth]{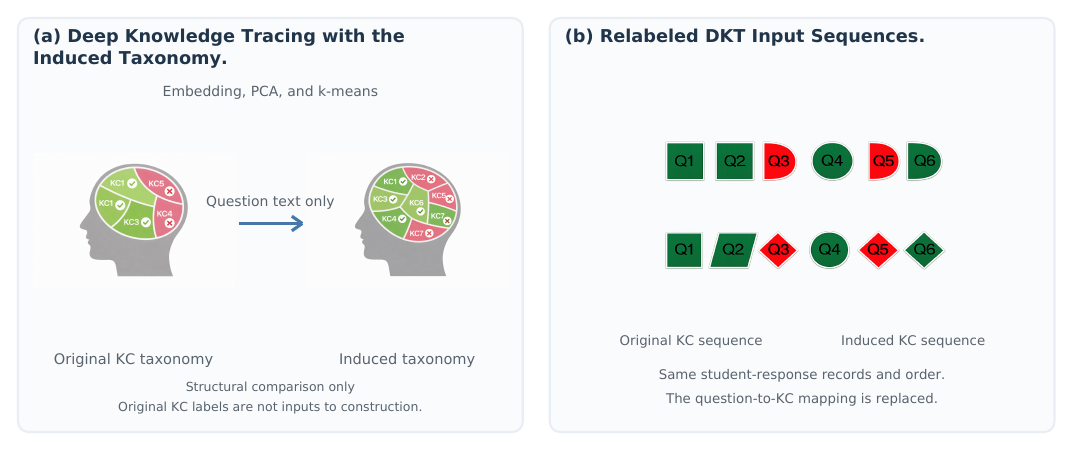}
\caption{Knowledge tracing with an induced taxonomy and sequence relabeling.
Panel (a) conceptually illustrates DKT tracking induced KCs; the illustrated
mastery states are conceptual and do not represent measured student knowledge.
Panel (b) shows the actual relabeling procedure for DKT evaluation: the original
KC label in each interaction is replaced with the induced label while questions,
responses, sequence order, student splits, and evaluation remain unchanged.}
\label{fig:pipeline}
\end{figure*}

Figure~\ref{fig:pipeline} illustrates knowledge tracing with an induced taxonomy
and the corresponding relabeling used for DKT evaluation. We first derive
a mapping from questions to categories that has one level, using question text
alone. We then replace the KC mapping supplied with the dataset with the induced
mapping in DKT while keeping the records of student answers and their order
unchanged.

\subsection{Data and Building the Taxonomies}

We use ASSISTments 2012 and 2017, two releases from a widely used benchmark
family in knowledge tracing~\cite{fengAddressingAssessmentChallenge2009,
priharExploringCommonTrends2022,abdelrahmanKnowledgeTracingSurvey2023}, because
their student interactions can be linked to question text from the ASSISTments
Problem Bodies (PB) resource~\cite{assistmentsProblemBodies}. The full 2012
release contains 179,999 questions and 265 KCs, of which 32,792 questions and
162 KCs belong to the PB text-linked subset. The full 2017 release contains
3,162 questions and 102 KCs, of which 638 questions and 68 KCs belong to the PB
text-linked subset. These PB subsets are the text-bearing questions and KCs
used for our text-based analyses. The largest category in
the original 2017 taxonomy, noskill, is a non-substantive placeholder covering
142 of the 638 questions (22\%). We exclude ASSISTments
2009 because its archived
release is sensitive to preprocessing choices~\cite{xiongGoingDeeperDKT2016}.
We refer to the dataset-provided question-to-KC mapping in each release as the
original taxonomy. Table~\ref{tab:experimental_design} summarizes the dataset
sizes, the category counts tested, and the PCA and \(K\) setting reported for each
dataset in the taxonomy comparison and DKT results.

\begin{table*}[t]
\centering
\small
\setlength{\tabcolsep}{5.2pt}
\begin{tabular}{lrrrrrr}
\toprule
Dataset & \shortstack{Total\\questions} & \shortstack{PB\\questions} &
\shortstack{Total\\KCs} & \shortstack{PB\\KCs} & Students & Responses \\
\midrule
ASSISTments 2012 & 179,999 & 32,792 & 265 & 162 & 23,845 & 978,271 \\
ASSISTments 2017 & 3,162 & 638 & 102 & 68 & 1,703 & 237,092 \\
\bottomrule
\end{tabular}
\caption{Dataset and taxonomy summary. Total counts describe the full releases;
PB counts describe the Problem Bodies text-linked question/KC subsets used for
text-based analyses. Student and response counts describe the text-linked subset
used for DKT evaluation. For both datasets, we construct induced taxonomies
using outputs from OpenAI's \texttt{text-embedding-3-small} and
\texttt{text-embedding-3-large}, shortened through the API's \texttt{dimensions}
parameter to 768 and 1,536 dimensions, respectively~\cite{openaiVectorEmbeddings};
we use the PCA dimensions \(\{10,20,40,60,80,100,120,160\}\), with \(K\) values
\(\{50,100,150,200,250,300\}\) for 2012 and \(\{20,40,60,80,100,150\}\) for
2017. PCA and \(k\)-means use a random seed of 42; \(k\)-means uses 10
initializations and at most 300 iterations. The reported (PCA, \(K\)) settings
in Table~\ref{tab:main_summary} are (100, 300) for 2012 and (10, 100) for 2017.
The PCA dimension and \(K\) are fixed before each DKT Bayesian search and are
not selected by it.}
\label{tab:experimental_design}
\end{table*}

We convert the HTML in each \texttt{problem\_body} to plain text, retain image
descriptions, remove scripts and style information, and normalize whitespace.
We construct each induced taxonomy from these cleaned question texts. The
original KC labels and student responses are not used during construction.
Two models embed each question, PCA reduces the vectors, and \(k\)-means assigns
each question to the nearest cluster center. We evaluate eight PCA dimensions and
six values of \(K\), the number of induced categories, for each dataset. The
tested grids (2012: 50--300; 2017: 20--150) span values below, near, and above
the original PB KC counts of 162 and 68, respectively. Each
taxonomy assigns one label per question and has one level, not a hierarchy. The
taxonomy is constructed from the complete set of linked question texts before
splitting students. Following the standard evaluation for knowledge tracing
implemented in pyKT~\cite{liuPyKTPythonLibrary2023}, we split the data by student
and evaluate students held out from training. The test set therefore contains
students not seen during training rather than previously unseen questions.

Appendix~\ref{sec:app:examples} documents the permitted examples of question text,
and Appendix~\ref{sec:app:dkt_settings} gives the exact embedding models, PCA
dimensions, values of \(K\), clustering settings, and procedure used to select
DKT settings.

Problem Bodies were accessed under the ASSISTments terms of use for the data. The
embedding inputs consisted only of cleaned text from the
\texttt{problem\_body} field and excluded grade information, original KC labels,
and records of student answers. Because the access terms prohibit redistribution
of the questions, neither the paper nor \PaperRepositoryReference{} reproduces
the original question text.

\subsection{Taxonomy Comparison}

We compare questions that have both an original and an induced label using the
measures summarized in Appendix Table~\ref{tab:metric-definitions}. Homogeneity
measures whether each induced category primarily contains questions from one
original category, while completeness measures whether each original category
remains concentrated within one induced category. The remaining measures ask how
similarly the two taxonomies group questions, whether pairs of questions stay
together, and how different the groupings are overall. Because some scores change
simply when the induced taxonomy uses more categories, we interpret them together
rather than treating any one score as a complete evaluation. We report induced
category sizes separately. For the reported 2017 DKT setting,
Appendix Figure~\ref{fig:app-contingency} shows how the 25 largest original
categories are distributed across the induced taxonomy. Together, these measures
describe how categories stay together, split, or mix across the two taxonomies.
Educational usefulness requires separate evidence.

\subsection{DKT Evaluation}
\label{sec:dkt}

We use DKT because it can use a student's performance on one KC when predicting
later answers~\cite{piech_deep_2015}. KC labels are part of each sequence given to
DKT, so replacing the original labels tests how DKT performs with the induced
taxonomy as a whole. We keep every question, response, and sequence position
fixed and replace the original mapping from questions to KCs with the
corresponding induced mapping. pyKT maps the resulting labels to consecutive
concept identifiers~\cite{liuPyKTPythonLibrary2023}. Varying \(K\) changes the
number of labels and can also change the number of model
parameters tied to those labels. Figure~\ref{fig:pipeline}b illustrates this
relabeling.

We run a separate Bayesian search for the original taxonomy and for each induced
taxonomy defined by an embedding model, PCA dimension, and \(K\).
Within each dataset, all searches use the same DKT model family, candidate
settings, number of trials, student split, training procedure, and selection
rule. We reserve 20\% of students for final testing, giving an approximate
64/16/20 training, validation, and test split. Within each search, we select the
completed run with the highest validation AUC and report its test AUC. The
embedding model, PCA dimension, and \(K\) are fixed before the search. Testing
both embedding models across eight PCA dimensions and six values of \(K\) yields
96 induced taxonomies per dataset. Table~\ref{tab:main_summary} reports the
highest test AUC across the induced taxonomies and the embedding model, PCA
dimension, and \(K\) that produced it; Figure~\ref{fig:auc_grid} shows the
grid of PCA dimensions and \(K\) values for that embedding model.
Appendix~\ref{sec:app:dkt_settings}
gives the exact search spaces and training settings.

To test whether closer agreement between the taxonomies is related to higher AUC,
we analyze the 48 settings of PCA dimension and \(K\) in each dataset. At each
setting, we average each structural score across the two embedding models and
separately average the two test AUC values. We
compute Spearman correlations, then repeat the analysis after removing rank
patterns linked to PCA dimension and \(K\). We apply the Benjamini--Hochberg
correction separately to each set of 15 comparisons~\cite{
benjaminiControllingFalseDiscovery1995}.
Appendix~\ref{sec:app:auc_relation} reports the full procedure and coefficients;
the analysis shows associations, not causal effects.

\section{Results}

\subsection{RQ1: Comparing the Original and Induced Taxonomies}

Table~\ref{tab:main_summary} shows much stronger agreement for
ASSISTments 2012 than for ASSISTments 2017 at the two reported settings. For
2012, homogeneity, completeness, and normalized mutual information (NMI) are
.930, .717, and .809; for 2017, they are .681, .509, and .582. Neither setting
yields the highest adjusted Rand index (ARI) or NMI across all
tested settings, showing that the settings reported for DKT are not simply those
with the closest match. The full comparison results
appear in Appendix~\ref{sec:app:structural}; Appendix
Table~\ref{tab:dkt-anchor-metrics} gives additional comparison scores at these
two settings. Appendix Figure~\ref{fig:app-contingency} shows how the 142
questions in noskill are distributed across the induced taxonomy.

\begin{figure*}[t]
\centering
\includegraphics[width=.98\textwidth]{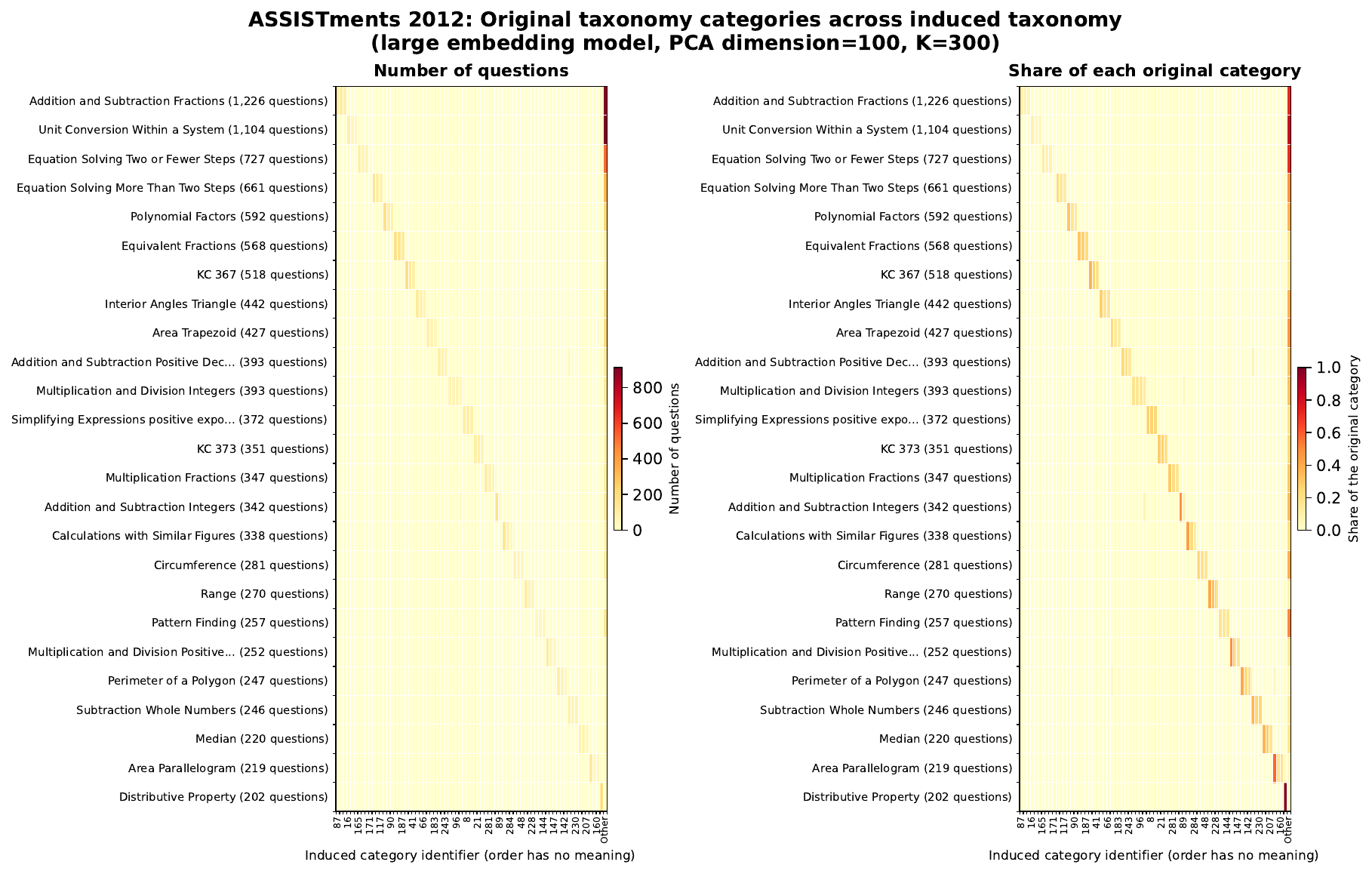}
\caption{Distribution of questions from the 25 largest categories in the 2012
original taxonomy across the corresponding induced taxonomy. The figure uses
\texttt{text-embedding-3-large}, PCA\(=100\), and \(K=300\). The left heatmap
shows question counts and the right heatmap shows the share of each original
category in each induced category. We include the three induced categories
containing the most questions from each original category and combine the rest
as \textit{Other}. The identifiers for induced categories are arbitrary labels
and do not indicate an order.}
\label{fig:taxonomy_distribution}
\end{figure*}

Figure~\ref{fig:taxonomy_distribution} shows that questions from large original
2012 categories are often distributed across several induced categories. These
divisions may reflect finer mathematical distinctions, recurring wording or
templates, or the effect of using more categories. This analysis does not show
whether these divisions matter for teaching.

\begin{table*}[t]
\centering
\scriptsize
\setlength{\tabcolsep}{2.8pt}
\begin{tabular}{llrrrrrrrrr}
\toprule
Dataset & Embedding & PCA & \(K\) & Homogeneity & Completeness & NMI & ARI &
\shortstack{AUC with\\original taxonomy} &
\shortstack{Highest AUC with\\induced taxonomy} & \(\Delta\)AUC \\
\midrule
ASSISTments 2012 & 3-large & 100 & 300 & .930 & .717 & .809 & .293 & .7367 & .7841 & +.0474 \\
ASSISTments 2017 & 3-large & 10 & 100 & .681 & .509 & .582 & .066 & .7131 & .7403 & +.0272 \\
\bottomrule
\end{tabular}
\caption{Taxonomy comparison and DKT results at the reported settings. Scores
for taxonomy agreement are averaged across the two embedding models at each
listed PCA dimension and \(K\). The embedding column identifies the model used
for the induced taxonomy with the highest observed test AUC. The induced value
is the highest observed test AUC across the tested combinations of embedding
model, PCA dimension, and \(K\). \(\Delta\)AUC is induced minus original.}
\label{tab:main_summary}
\end{table*}

\subsection{RQ2: Effect of the Number of Categories}

\begin{figure*}[t]
\centering
\includegraphics[width=.92\textwidth]{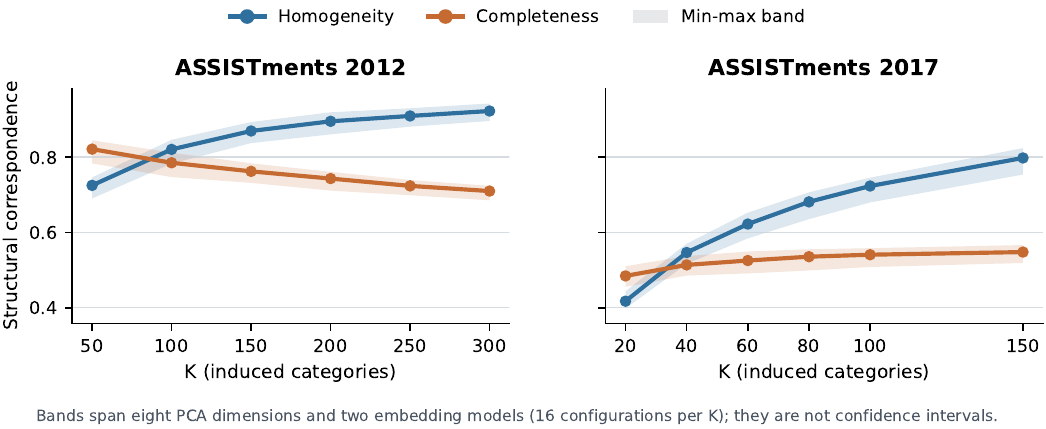}
\caption{Mean homogeneity and completeness as the induced taxonomy uses more
categories. At each \(K\), the lines show averages over the eight PCA dimensions
and two embedding models; bands show the full range across those 16 settings and
are not confidence intervals. In 2012, homogeneity generally increases while
completeness decreases. In 2017, both generally increase, although inverse
purity and pairwise recall decline over the same \(K\) range. Appendix
Figure~\ref{fig:app-hc} gives the full results for each embedding model and PCA
dimension.}
\label{fig:granularity_summary}
\end{figure*}

Figure~\ref{fig:granularity_summary} shows how homogeneity and completeness change
as \(K\) increases. In ASSISTments 2012, homogeneity generally rises while
completeness falls: induced categories become more concentrated, while each
original KC is distributed across more induced categories. In ASSISTments 2017,
both averages generally rise over the tested range, but inverse purity and
pairwise recall decline, as shown in Appendix
Table~\ref{tab:granularity-endpoints}. The measures therefore capture different
effects of using more categories and do not indicate improvement in every aspect
of the match.

\subsection{RQ3: Taxonomies and DKT Performance}

\begin{figure*}[t]
\centering
\begin{minipage}[t]{.49\textwidth}
\centering
\includegraphics[width=\linewidth]{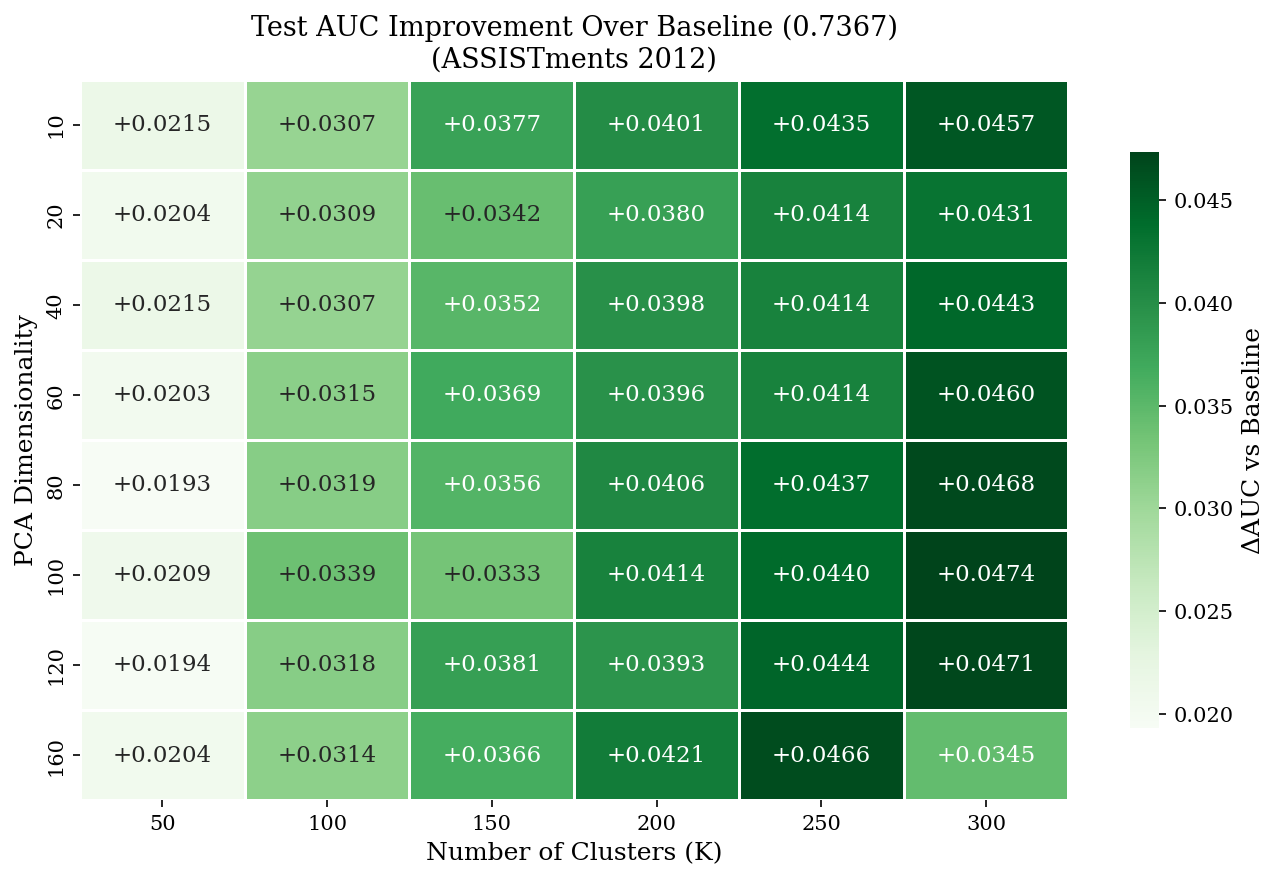}
\end{minipage}\hfill
\begin{minipage}[t]{.49\textwidth}
\centering
\includegraphics[width=\linewidth]{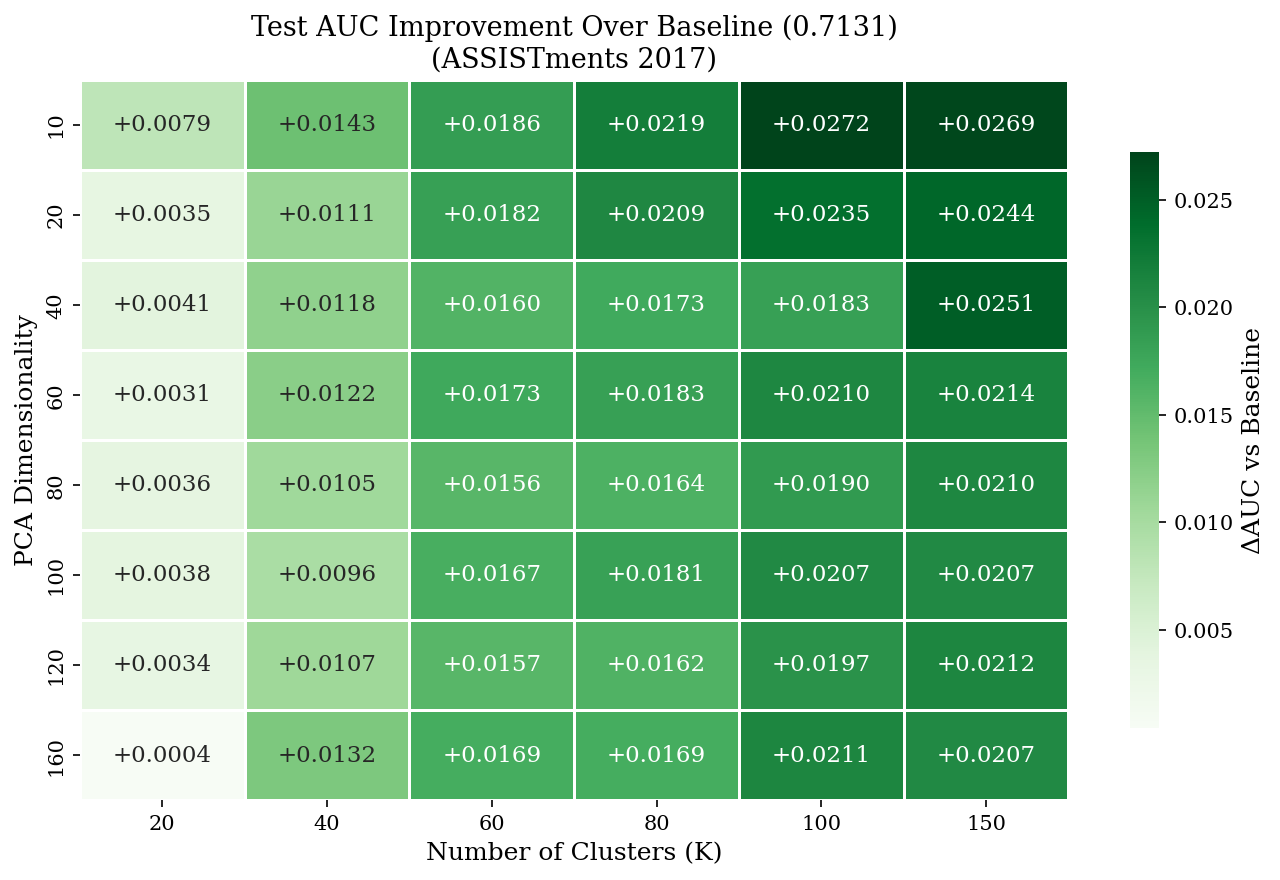}
\end{minipage}
\caption{Change in test AUC when using the induced taxonomy instead of the
original taxonomy. The figure uses \texttt{text-embedding-3-large}, which produced the
highest observed test AUC in both datasets. Each cell shows the change in test
AUC for one PCA and \(K\) setting. The test AUC at every displayed setting
exceeds the baseline with the original taxonomy. We also test
\texttt{text-embedding-3-small}, giving 96 induced taxonomies per dataset in
total.}
\label{fig:auc_grid}
\end{figure*}

Table~\ref{tab:main_summary} reports that the highest test AUC among the induced
taxonomies exceeds the AUC with the original taxonomy in both datasets. For
ASSISTments 2012, AUC rises from .7367 with the original taxonomy to .7841, a
gain of .0474. For ASSISTments
2017, it rises from .7131 to .7403, a gain of .0272.
Figure~\ref{fig:auc_grid} shows that, for \texttt{text-embedding-3-large}, test
AUC exceeds the original-taxonomy baseline at all 48 PCA/\(K\) grid settings
in each dataset, including settings with fewer categories than the corresponding
original taxonomy within the PB subset.

Before adjustment, many comparison measures correlate strongly with AUC because
both vary with PCA dimension and \(K\). After accounting for those patterns and
correcting for 15 comparisons, none of the 2012 correlations remains
statistically significant. In 2017, several measures of closer agreement are
associated with lower AUC, while a larger variation of information score is
associated with higher AUC. Closer agreement is therefore not consistently
associated with higher DKT AUC across the tested settings. These correlations
do not show that differences from the original taxonomy cause the gains.
Appendix~\ref{sec:app:auc_relation} reports the full coefficients in
Table~\ref{tab:auc_correlations} and Figure~\ref{fig:auc_correlations}.

\section{Discussion}

The original and induced KC taxonomies organize the same questions using
different evidence. The stronger agreement in ASSISTments 2012 suggests that
patterns in question text align more closely with the original KC structure
there than in ASSISTments 2017. In both datasets, however, original categories
are split across several induced categories. These splits may reflect
mathematical content, recurring wording, or the use of more induced categories;
the current analysis cannot determine which explanation is correct.

The 2017 original taxonomy's largest category, \textit{noskill}, is a
non-substantive placeholder covering 142 of 638 questions (22\%). Because
\textit{noskill} groups questions by the absence of an assigned KC rather than
shared content, a taxonomy constructed from question content has no reason to
reproduce it as a single category. Appendix
Figure~\ref{fig:app-contingency} shows the induced taxonomy distributing these
questions across several categories. The original-taxonomy baseline retains
\textit{noskill}, so the reported AUC difference compares DKT with this
placeholder against DKT with the complete induced mapping. This redistribution
may contribute to both the lower correspondence and the higher AUC observed in
2017, although the experiment does not isolate its contribution from the other
changes in the mapping.

The DKT results address a different question: how well each taxonomy supports
prediction. The induced taxonomy with the highest observed test AUC improves
test AUC by .0474 in 2012 and .0272 in 2017, and Figure~\ref{fig:auc_grid}
shows improvement across all 48
PCA and \(K\) settings for \texttt{text-embedding-3-large} in both datasets.
These results evaluate each induced taxonomy as a whole, including its grouping
of questions, number of categories, label frequencies, and model parameters tied
to its labels. They do not show which property drives the gain,
and AUC measures prediction rather than educational value. After accounting for
PCA dimension and \(K\), closer agreement with the original taxonomy is not
consistently associated with higher AUC. The negative correlations in 2017
therefore show that lower agreement is sometimes associated with higher AUC,
not that disagreement is inherently beneficial.

The induced taxonomies are therefore best treated as complements to systems
developed by experts. Their similarities and differences can highlight categories
or boundaries for expert review. Determining whether these distinctions matter
for teaching or reflect how students reason requires evidence from educators,
student behavior, and solution steps.

A comparison with biological taxonomy helps clarify the idea. Biological
categories are sometimes reconsidered using genetic evidence alongside visible
traits~\cite{sanMauroMolecularSystematics2010}. In the same way, text embeddings
provide another source of evidence that may support some expert categories and
question others. The comparison has a clear limit: similar question text is not
automatically a better basis for curriculum design. Educational categories are
useful when they support teaching, assessment, or models of learning.

\section{Future Work}

Future work should extend the analysis to datasets that combine student
interactions, question text, and more detailed KC structures. XES3G5M contains
5.5 million responses, assigns multiple KCs to some questions, and links broader
skills to narrower ones. DBE-KT22 provides question text and student responses
from a university database course, assigns up to four KCs per question, and
includes relationships among KCs~\cite{
liuXES3G5M2023,abdelrahmanDBEKT222022}. These datasets would support induced
taxonomies with multiple labels per question or hierarchical relations among
categories. The current ASSISTments datasets would remain useful for evaluating
models that explicitly support these richer structures. The induced taxonomies could
also be evaluated in other tasks that use KCs, including measuring progress with
practice, identifying students' relative strengths and weaknesses, and
recommending the next practice question~\cite{wangNeuralCognitiveDiagnosis2020,
cenLearningFactorsAnalysis2006,aiConceptAwareRecommendation2019}.

\section{Conclusion}

We use embeddings of question text to build educational taxonomies and compare
them with the original KC taxonomies in ASSISTments 2012 and 2017. The induced
taxonomies preserve much, but not all, of the original structure, with closer
agreement in 2012 and consistent changes as the number of induced categories
increases. In DKT, the induced taxonomy with the highest observed test AUC
improves test AUC by .0474 in 2012 and .0272 in 2017,
while closer agreement with the original taxonomy is not consistently
associated with higher AUC across the tested settings. The central result
therefore separates two questions: similarity to a taxonomy developed by experts and usefulness for
prediction. Taxonomies built from embeddings offer a promising alternative for
student models, but their educational value still requires review by experts and
evidence from students.

\FloatBarrier
\label{main-content-end}
\section{Limitations}

Our analysis covers only ASSISTments 2012 and 2017 questions with linked text.
Both datasets come from the same learning platform, so the findings may not
extend to other platforms, courses, schools, languages, or educational settings.
We include only responses tied to questions with text.

The procedure for building the taxonomy assigns each question to one induced
category. It does not allow several KCs per question, organize KCs from broad to
narrow, or model student reasoning. The induced taxonomy may organize questions
by wording, templates, formatting, answer type, or source rather than by the
mathematics or knowledge required to answer them.

The comparison measures show how closely the two taxonomies group the same
questions rather than their educational value. Purity and homogeneity can rise
simply with the number of categories, and overall scores do not identify which
specific splits or mergers matter for instruction. We do not analyze solution
steps, inspect individual category boundaries, or ask educators to judge the
categories, so we cannot determine the educational value of those differences.

The experiments use two models for embedding text, one clustering method, one
clustering seed, and a fixed set of PCA dimensions and values of \(K\). The
observed results may therefore depend on the tested models, clustering procedure,
category count, question templates, or distribution of cluster sizes. We use PCA
as the only dimensionality-reduction method; alternative reduction methods may
produce different induced taxonomies. We do not directly benchmark against prior
KC-discovery methods such as SMART or KCluster.

We evaluate one DKT architecture. Although all taxonomies use the same records,
student split, training procedure, search protocol, and trial
budget, we tune DKT separately for each taxonomy, so the selected
hyperparameters may differ. Changing \(K\) also changes the number of KC labels
and can change the number of model parameters tied to those labels. Without
random-label baselines matched on the number and sizes of categories, the
experiment cannot separate grouping by meaning from the effects of category
count, how often each label appears, or model size.

The reported induced AUC is the highest result across all tested combinations of
embedding model, PCA dimension, and \(K\). Because this combination is identified
using test AUC, the reported gain summarizes the best tested result rather than
a combination chosen before looking at test performance.
Figure~\ref{fig:auc_grid} shows the results across the full grid. For each
combination, the Bayesian search chooses DKT settings using validation AUC and
does not choose the embedding model, PCA dimension, or \(K\). Because students
rather than questions are held out, test AUC measures performance for new
students on questions used to build the taxonomy, not entirely new questions.
Larger \(K\) may also make induced labels behave more like coarse question
identifiers, which could contribute to higher AUC.
Finally, the analysis of taxonomy agreement and AUC shows an association and does
not establish that changing agreement causes the AUC gains.

\section*{Acknowledgments}

We thank the ASSISTments Project at Worcester Polytechnic Institute for
providing the Problem Bodies data, which allowed us to link records of student
responses with question text~\cite{assistmentsProblemBodies}.

\bibliography{references}

\appendix
\raggedbottom
\section{Question Text Inputs and Examples}
\label{sec:app:examples}

\paragraph{Question examples.}
The embedding models use only the cleaned text from the
\texttt{problem\_body} field in \texttt{questions.csv}; grade information and
the original category labels are excluded. The grade labels range from
elementary school through high school mathematics and do not necessarily match
a student's enrolled grade. Although Grades 6--8 usually correspond to U.S. middle school,
the data cover a broader range. The Grade 4 and Grade 6 examples also
illustrate that converting web pages to plain text can remove mathematical
layout and diagrams. Images are not provided to the models. Because the access
terms for Problem Bodies prohibit quoting the questions, we describe
representative examples instead~\cite{assistmentsProblemBodies}.
\begin{itemize}
    \item \textbf{ASSISTments 2012, Grade 4:} converting an improper fraction.
    \item \textbf{ASSISTments 2012, Grade 6:} finding a square's area from a
    diagram.
    \item \textbf{ASSISTments 2017, Grade 7:} finding a percentage of a whole.
    \item \textbf{ASSISTments 2017, Grade 8:} using the formula for a cone's
    volume.
    \item \textbf{ASSISTments 2012, High School Algebra:} comparing the
    degrees of polynomial terms.
\end{itemize}

\section{Taxonomy and DKT Settings}
\label{sec:app:dkt_settings}

By default, OpenAI's \texttt{text-embedding-3-small} and
\texttt{text-embedding-3-large} produce embeddings with 1,536 and 3,072
dimensions, respectively. We use the API's \texttt{dimensions} parameter to
request outputs with 768 and 1,536 dimensions,
respectively~\cite{openaiVectorEmbeddings}. Let \(e_i\) denote the embedding for
question \(i\).
PCA reduces it to \(z_i=P(e_i)\), and \(k\)-means assigns \(z_i\) to the nearest
centroid:
\[
c_i=\arg\min_{j\in\{1,\dots,K\}}\lVert z_i-\mu_j\rVert_2^2.
\]
Here, \(c_i\) is the label assigned by the induced taxonomy, \(K\) is the number
of induced categories, and \(\mu_j\) is the centroid of cluster \(j\). We test
PCA dimensions
\(\{10,20,40,60,80,100,120,160\}\). We test values of \(K\)
\(\{50,100,150,200,250,300\}\) for 2012 and
\(\{20,40,60,80,100,150\}\) for 2017. PCA and \(k\)-means use
a random seed of 42.
For each setting, \(k\)-means uses 10 initializations and at most 300 iterations
(\texttt{n\_init=10}, \texttt{max\_iter=300}).

For response \(t\), \(q_t\) denotes the question, \(s_t\) its original KC label,
\(r_t\) whether the response is correct, and \(c(q_t)\) the label assigned by the
induced taxonomy.
We apply the relabeling
\[
(q_t,s_t,r_t)\mapsto(q_t,c(q_t),r_t).
\]
For 2012, we write induced labels to \texttt{skill\_id}. For 2017, we write them as
text in \texttt{skill}. pyKT then renumbers them as consecutive concept IDs.

For the baseline that uses the original taxonomy and for every induced taxonomy,
we run a separate Weights \& Biases Bayesian search that maximizes validation
AUC. Each induced taxonomy is defined by its dataset, embedding model, PCA
dimension, and \(K\). Within each dataset, all searches use the same candidate settings, number
of trials, assignment of students to training, validation, and testing, training
procedure, and selection rule.

For 2012, the search varies \texttt{emb\_size} over
\(\{64,128,256,512\}\), \texttt{learning\_rate} over
\(\{0.01,0.001,0.0007,0.0005\}\), \texttt{dropout} over
\(\{0.1,0.2,0.3,0.5\}\), and \texttt{seed} over \(\{42,84\}\). For 2017,
the search varies \texttt{emb\_size} over
\(\{64,96,128,192,256,384,512,768,1024\}\),
\texttt{learning\_rate} over
\(\{0.01,0.001,0.0007,0.0005,0.002,0.0002\}\), and
\texttt{dropout} over \(\{0.01,0.1,0.2,0.3,0.4,0.45,0.5\}\). The 2017 search
fixes \texttt{seed} at 42. For 2012, seed is one of the settings considered by
the search; it is not used to repeat each taxonomy.

Within each search, we select the completed run with the highest validation AUC
and report its test AUC; the embedding model, PCA dimension, and \(K\) remain
fixed outside the search.

Both the DKT/QID input format, which includes question and KC identifiers, and
fold 0 are fixed across all searches.
Training uses Adam with batch size 256, at most 200 epochs, and a maximum sequence
length of 200. The preprocessing code reserves 20\% of students for testing with
\texttt{random\_state=1024}; the remaining students are divided into five folds,
yielding the approximate 64/16/20 training/validation/test split. Training stops
after 30 epochs without a validation AUC improvement of at least .001.

\section{Measures for Comparing the Taxonomies}
\label{sec:app:structural}

Combining eight PCA dimensions with six values of \(K\) gives 48 settings
per dataset. We evaluate both embedding models at every setting, yielding 96
rows per dataset. Each analysis below states whether results for the two
embedding models are reported separately or averaged.

\begin{table}[!b]
\centering
\footnotesize
\setlength{\tabcolsep}{3pt}
\renewcommand{\arraystretch}{.94}
\begin{tabular}{p{.25\columnwidth}p{.68\columnwidth}}
\toprule
Measure & What the score measures \\
\midrule
ARI & Adjusted Rand index: agreement between pairs of questions, adjusted for
chance; higher is closer. \\
NMI / AMI & Normalized / adjusted mutual information: how much information the
two taxonomies share; AMI adjusts for chance; higher is closer. \\
Homogeneity & Whether each induced category mostly contains questions from one
original category; higher is closer. \\
Completeness & Whether questions from each original category mostly stay
together in one induced category; higher is closer. \\
V-measure & Harmonic mean of homogeneity and completeness; higher is closer. \\
Macro purity & Average share with the most common original label in each induced
category, weighting induced categories equally; higher is closer. \\
Global purity & Share of all questions with the most common original label in
their induced category, giving larger induced categories more weight; higher is
closer. \\
Global / macro inverse purity & How much of each original category remains in
its largest induced category, weighting categories by size or equally; higher is
closer. \\
FM & Fowlkes--Mallows: a combined score based on pairwise precision and recall;
higher is closer. \\
Pairwise P/R/\(F_1\) & Precision for pairs in the same induced category, recall
for pairs with the same original label, and their \(F_1\); higher is closer. \\
VI & Variation of information: how different the two taxonomies are based on
their information content; lower is closer. \\
\bottomrule
\end{tabular}
\caption{Measures used to compare the original and induced
taxonomies~\cite{rosenberg_v-measure_2007,vinh_information_2010,
meilaComparingClusterings2007,fowlkesMethodComparing1983}. Increasing \(K\) can
raise some scores and lower others, so the scores should be interpreted
together.}
\label{tab:metric-definitions}
\end{table}

\begin{table*}[t]
\centering
\small
\setlength{\tabcolsep}{5.0pt}
\begin{tabular}{lccccc}
\toprule
Dataset & Best ARI & Best AMI & Best V/NMI & Best FM & Lowest VI \\
\midrule
2012 & .490 (120, 50) & .812 (160, 150) & .830 (160, 150) &
.503 (120, 50) & 1.554 (160, 150) \\
2017 & .114 (100, 40) & .371 (100, 80) & .663 (120, 150) &
.165 (80, 20) & 2.789 (120, 150) \\
\bottomrule
\end{tabular}
\caption{Best agreement scores across the 48 settings, averaged across the two
embedding models. ARI is the adjusted Rand index;
AMI and NMI denote adjusted and normalized mutual information, respectively; FM is
Fowlkes--Mallows; and VI is variation of information. Parentheses give the PCA
dimension and \(K\). Lower VI means a closer match; higher values mean a closer
match for the other measures. NMI and V-measure have the same values here and
reach their highest values at the same setting.
Table~\ref{tab:granularity-endpoints} shows these and the purity measures at the
smallest and largest values of \(K\). Figure~\ref{fig:app-metrics} shows ARI and
NMI separately for each embedding model and dataset.}
\label{tab:structural-summary}
\end{table*}

\begin{table*}[t]
\centering
\scriptsize
\setlength{\tabcolsep}{2.7pt}
\resizebox{\textwidth}{!}{%
\begin{tabular}{lrrrrrrrrrrrr}
\toprule
Dataset (PCA, \(K\)) & Homog. & Complete. & V/NMI & Global pur. & Macro pur. &
Global inv. & Macro inv. & ARI & AMI & FM & Pair \(F_1\) & VI \\
\midrule
2012 \((100,300)\) & .930 & .717 & .809 & .888 & .898 & .406 & .617 &
.293 & .776 & .393 & .299 & 1.872 \\
2017 \((10,100)\) & .681 & .509 & .582 & .530 & .546 & .266 & .486 &
.066 & .272 & .123 & .083 & 3.290 \\
\bottomrule
\end{tabular}}
\caption{Structural scores at the two PCA/\(K\) settings used for the reported
DKT results. Each score is averaged across the two embedding models. ``Homog.''
means homogeneity, ``Complete.'' means completeness, ``pur.'' means purity, and
``inv.'' means inverse purity. Table~\ref{tab:metric-definitions} defines ARI,
AMI, and NMI; FM means Fowlkes--Mallows. VI uses natural logarithms and is the
only score for which lower values mean a closer match. V-measure and NMI have the
same values here.}
\label{tab:dkt-anchor-metrics}
\end{table*}

At the two reported combinations of PCA dimension and \(K\), the average spread in
cluster size across the two embedding models is .591 for 2012 and .447 for 2017.
This spread is the
standard deviation divided by the mean; a larger value means the cluster sizes
vary more. The average share of questions in the largest cluster is 1.28\% for
2012 and 2.51\% for 2017. Neither 2012 embedding model produces a cluster with
only one question;
each 2017 embedding model produces one such cluster. No model places more than 3\% of
questions in its largest cluster. All \(K\) clusters contain at least one question.
The mean cluster size is 62.45 for 2012 and 6.38 for 2017.

\begin{figure*}[t]
\centering
\includegraphics[width=.86\textwidth]{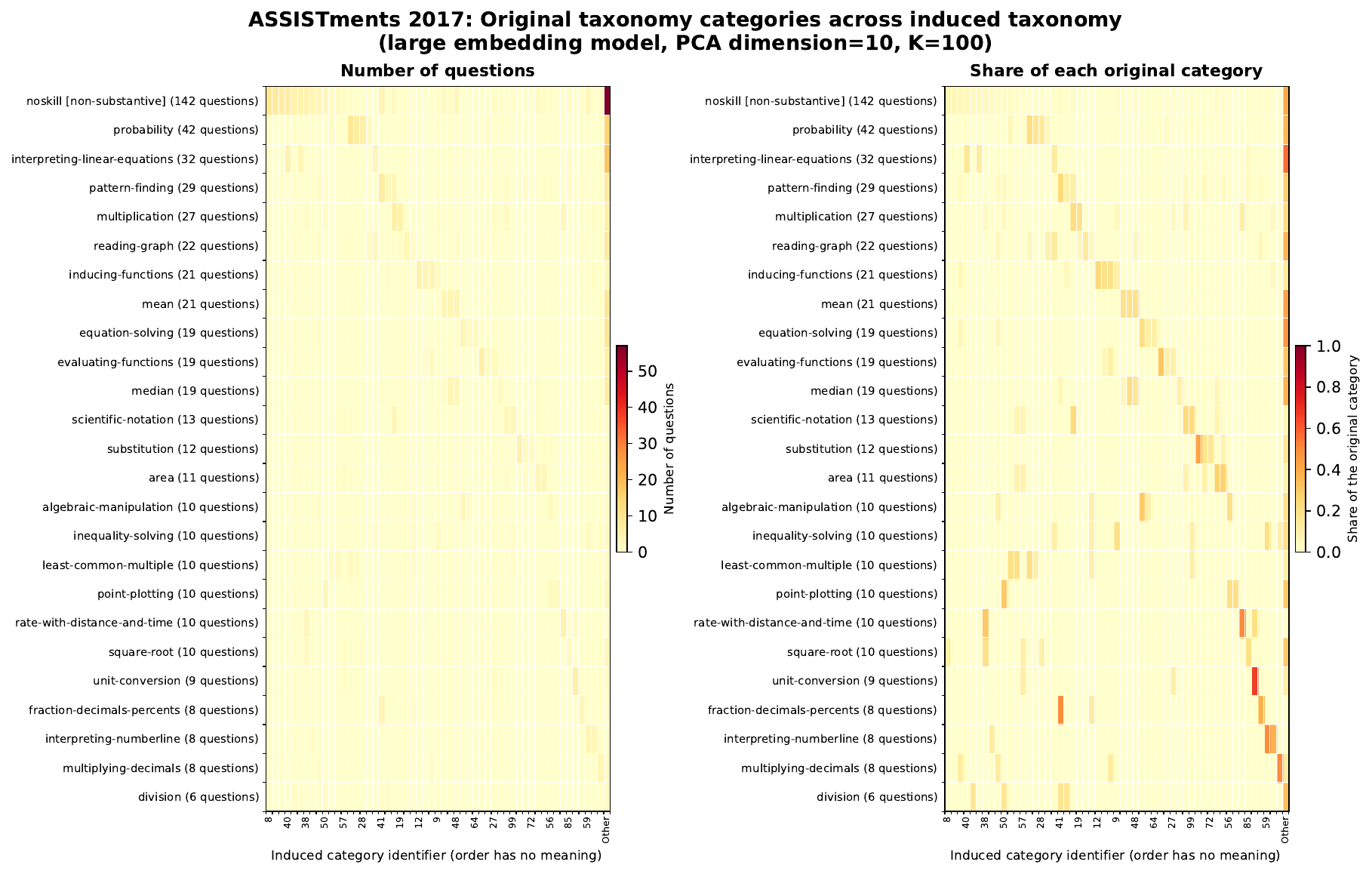}
\caption{Distribution of questions from the 25 largest categories in the 2017
original taxonomy across the corresponding induced taxonomy. The figure uses
\texttt{text-embedding-3-large}, PCA\(=10\), and \(K=100\). The left heatmap
shows question counts and the right heatmap shows the share of each original
category in each induced category. We include the three induced
categories containing the most questions from each original category and combine
the rest as \textit{Other}. Induced category identifiers are arbitrary labels and
do not indicate an order.}
\label{fig:app-contingency}
\end{figure*}

\begin{figure*}[t]
\centering
\textbf{(a) ASSISTments 2012}\par\smallskip
\includegraphics[width=.69\textwidth]{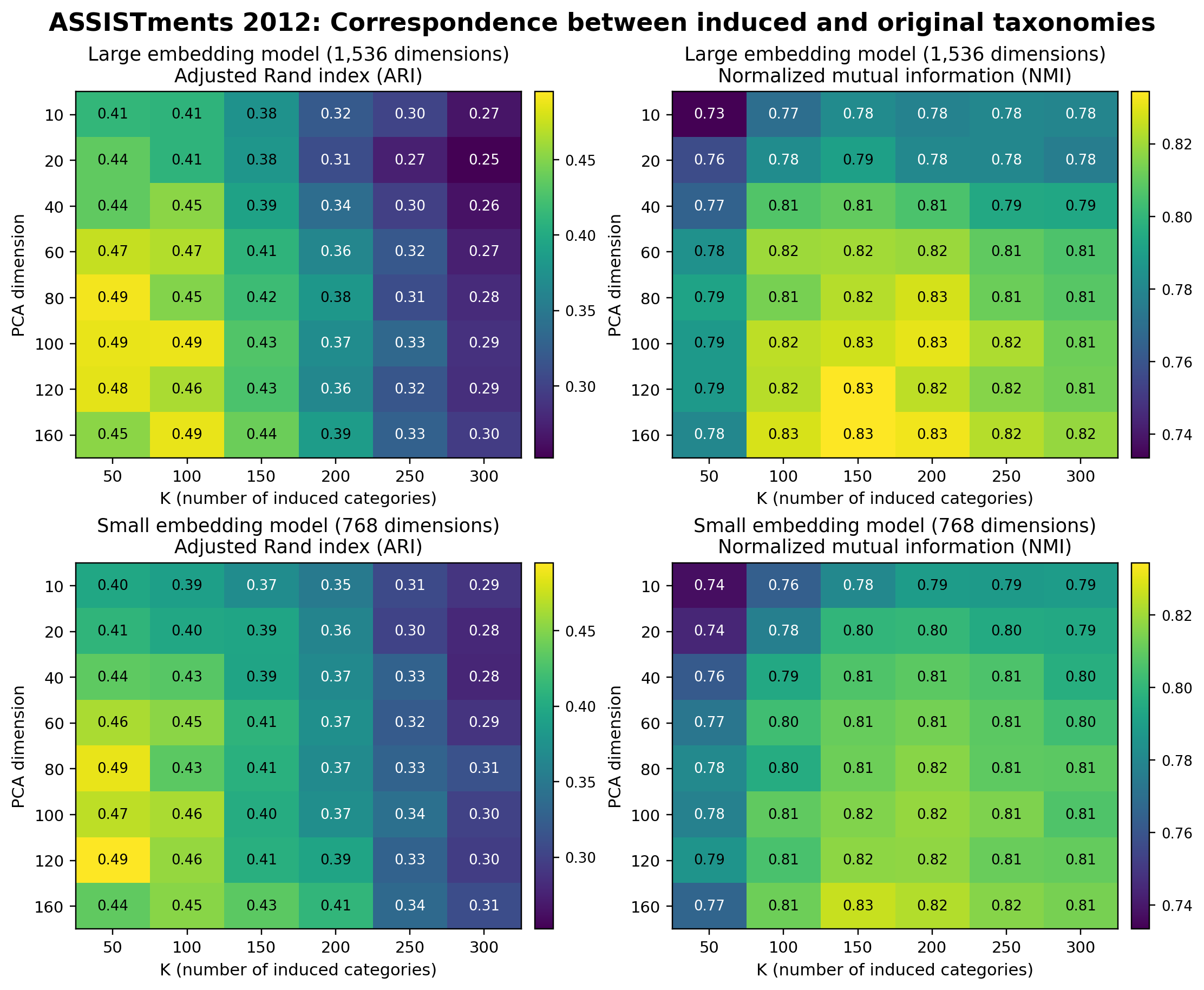}
\vspace{.5em}

\textbf{(b) ASSISTments 2017}\par\smallskip
\includegraphics[width=.69\textwidth]{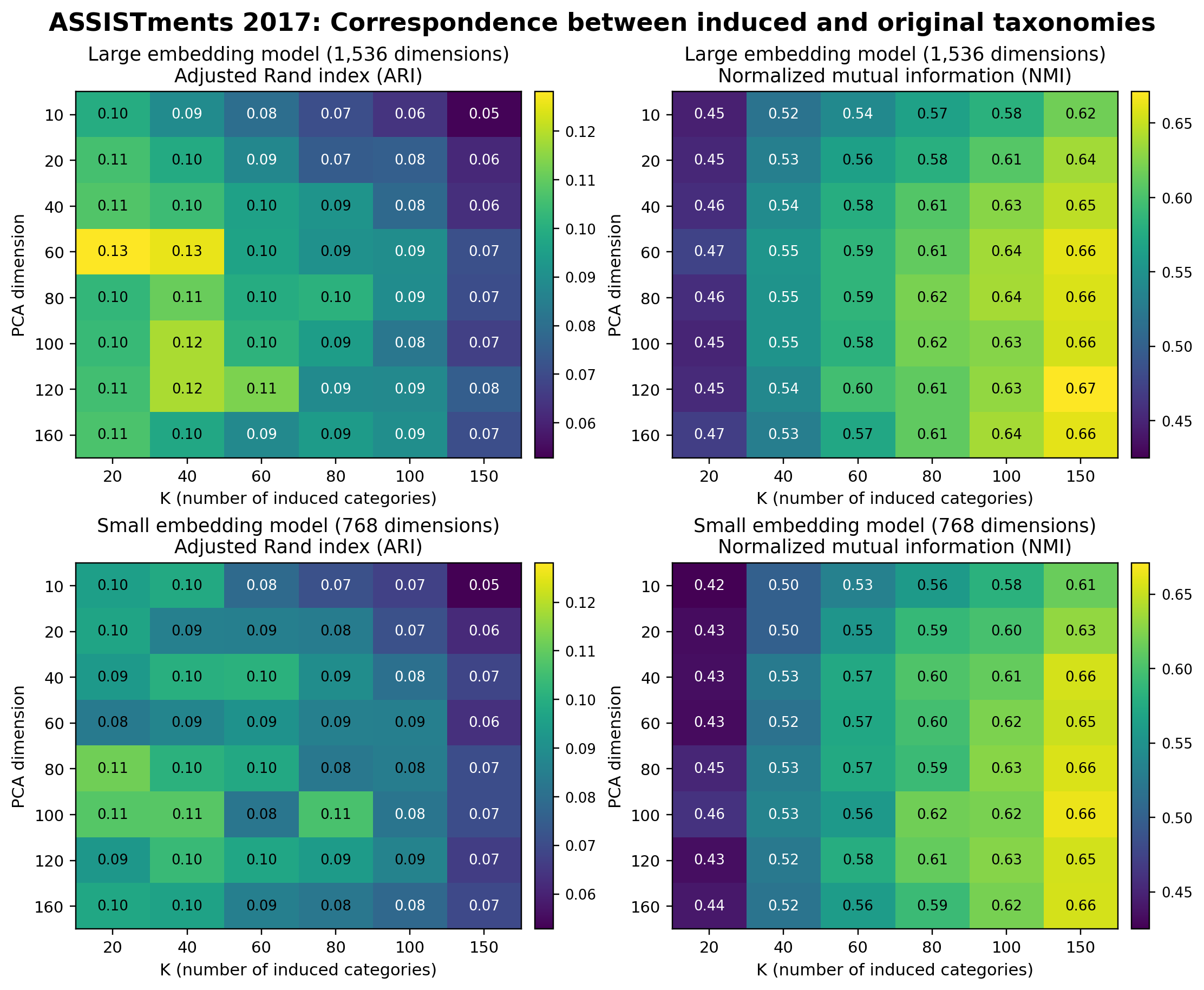}
\caption{ARI and NMI for every PCA dimension and value of \(K\), shown
separately for each embedding model in 2012 (a) and 2017 (b).
Table~\ref{tab:structural-summary} gives the highest values after averaging
scores across the two models. In 2017, ARI and NMI reach their highest values at
different settings, and, for each metric, each embedding model's highest value
differs from the highest value obtained after averaging scores across the two
models.}
\label{fig:app-metrics}
\end{figure*}

\begin{figure*}[t]
\centering
\textbf{(a) ASSISTments 2012}\par\smallskip
\includegraphics[width=.94\textwidth]{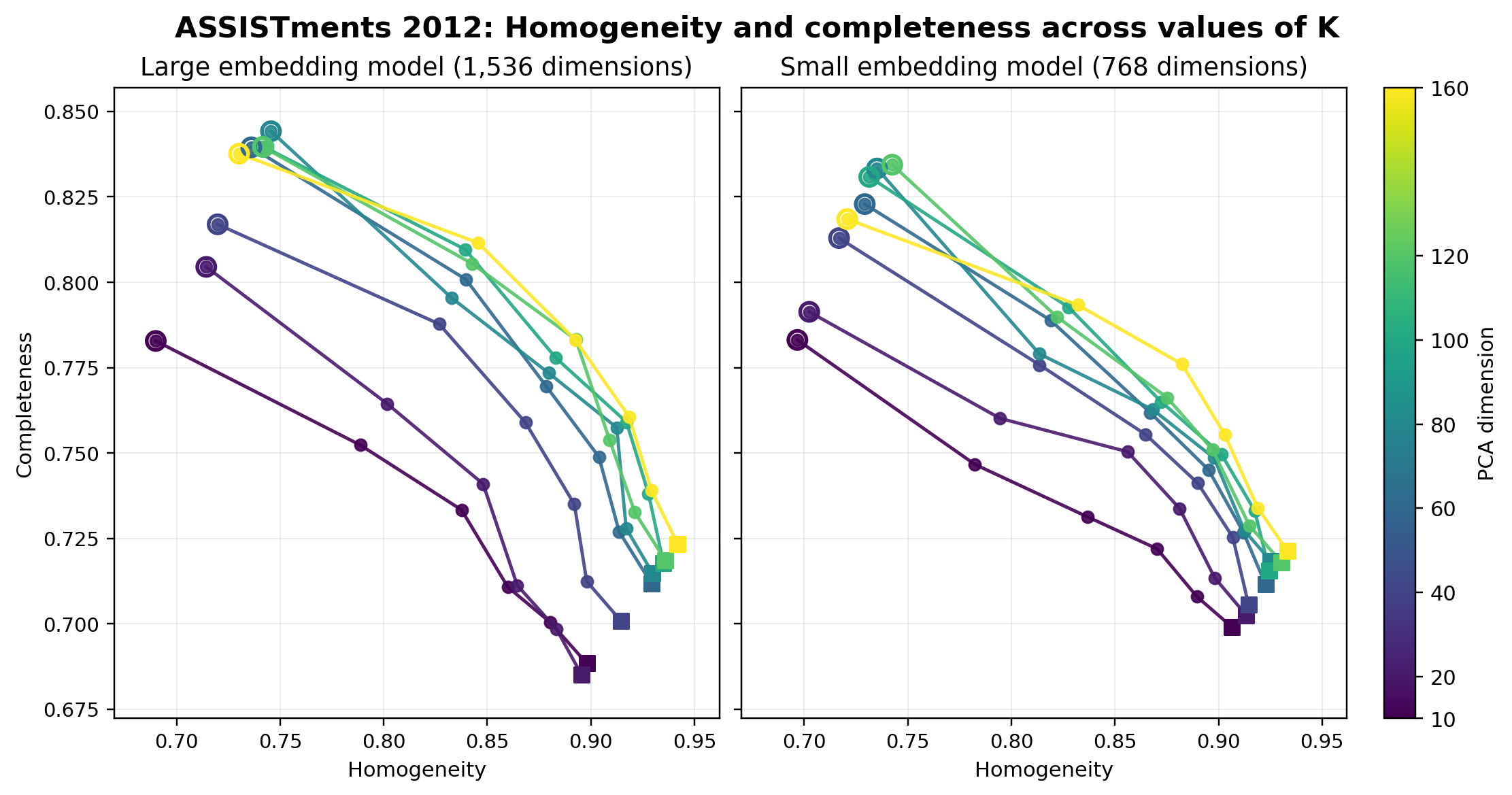}
\vspace{.5em}

\textbf{(b) ASSISTments 2017}\par\smallskip
\includegraphics[width=.94\textwidth]{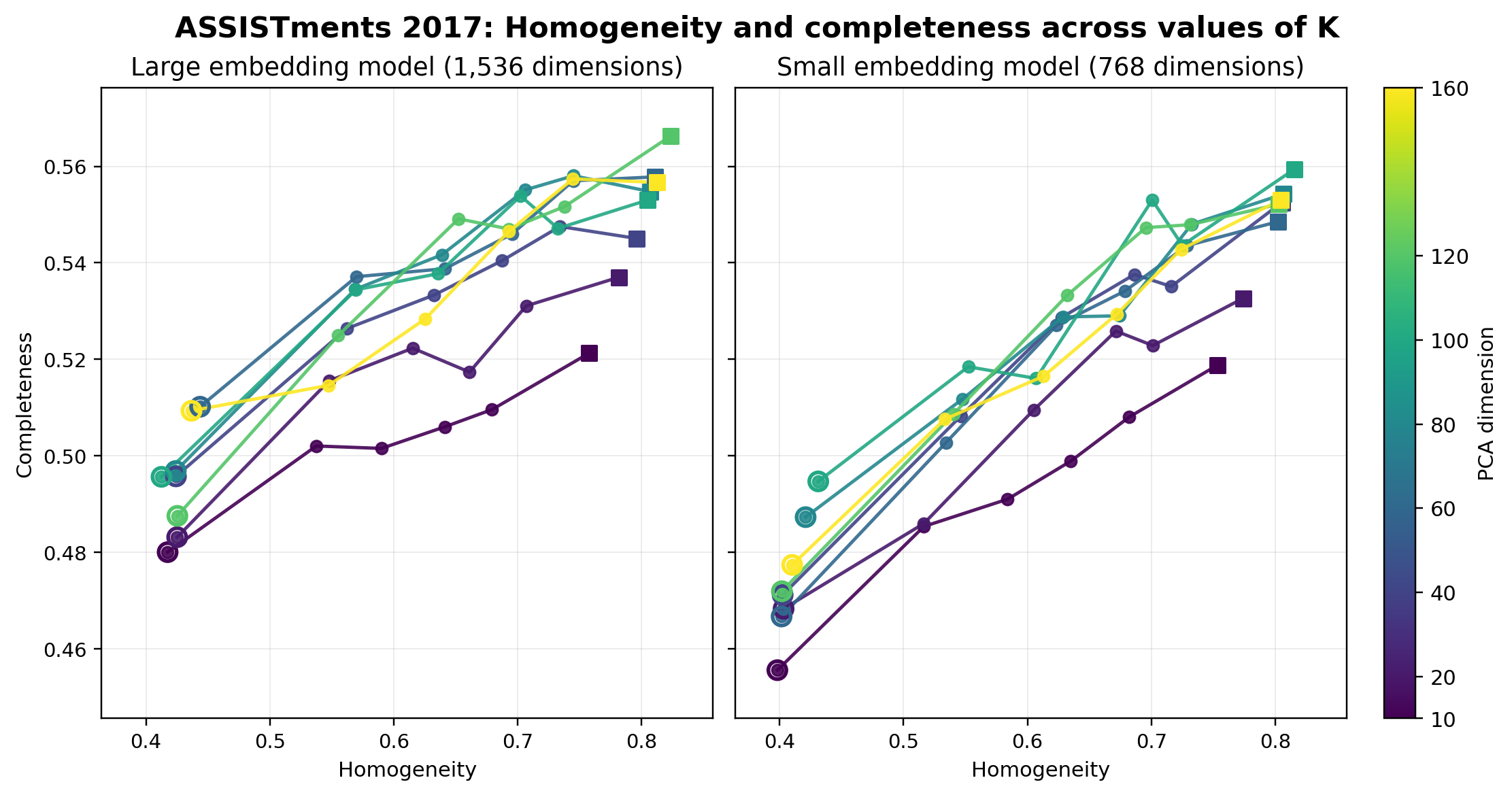}
\caption{Homogeneity and completeness across values of \(K\), shown separately
for each combination of embedding model and PCA dimension in 2012 (a) and 2017 (b). In 2012,
homogeneity generally rises while completeness falls. In 2017, both generally
rise across the tested settings, although global inverse purity and pairwise
recall fall from the smallest to the largest tested value of \(K\), as shown in
Table~\ref{tab:granularity-endpoints}. Outlined circles mark the smallest
tested \(K\), and squares mark the largest.}
\label{fig:app-hc}
\end{figure*}

\begin{table*}[t]
\centering
\scriptsize
\setlength{\tabcolsep}{3.0pt}
\begin{tabular}{lrrrrrrrrr}
\toprule
Dataset/\(K\) & Homog. & Complete. & Global purity & Global inverse &
Pair precision & Pair recall & VI & Size spread \\
\midrule
2012/50  & .725 & .821 & .572 & .717 & .426 & .516 & 1.854 & .522 \\
2012/300 & .922 & .710 & .880 & .392 & .839 & .176 & 1.948 & .575 \\
2017/20  & .417 & .484 & .345 & .439 & .175 & .146 & 3.460 & .418 \\
2017/150 & .797 & .548 & .674 & .270 & .442 & .042 & 2.901 & .476 \\
\bottomrule
\end{tabular}
\caption{Scores at the smallest and largest tested values of \(K\). Each value is
the simple average over eight PCA dimensions and both embedding models. In both
datasets, increasing \(K\) raises homogeneity, global purity, and pairwise
precision. It lowers global inverse purity and pairwise recall. In 2012,
completeness falls and VI rises. In 2017, completeness rises and VI falls.
``Homog.'' means homogeneity, ``Complete.'' means completeness, and VI means
variation of information. Size spread is the standard deviation of the induced
cluster sizes divided by their mean; a larger value means that the cluster sizes
differ more.}
\label{tab:granularity-endpoints}
\end{table*}
\FloatBarrier

\section{Relation Between Taxonomy Agreement and AUC}
\label{sec:app:auc_relation}

To examine the relationship between taxonomy agreement and AUC, we analyze 48
matched settings of PCA dimension and \(K\) per dataset. At each setting, we
average each structural score across the two embedding models and separately
average the two embedding-specific test AUC values. The 2017 AUC results include
three additional settings,
\((10,200)\), \((10,250)\), and \((10,300)\); we exclude them because the
taxonomy comparison has no matching scores.

We use Spearman correlation to test whether closer taxonomy agreement is
associated with higher test AUC. We then repeat the analysis after ranking both
measures and removing patterns linked to PCA dimension and \(K\).
Because we test 15 measures, we apply the Benjamini--Hochberg correction
separately by dataset and separately before and after this adjustment.

\begin{figure}[!b]
\centering
\includegraphics[width=.98\columnwidth]{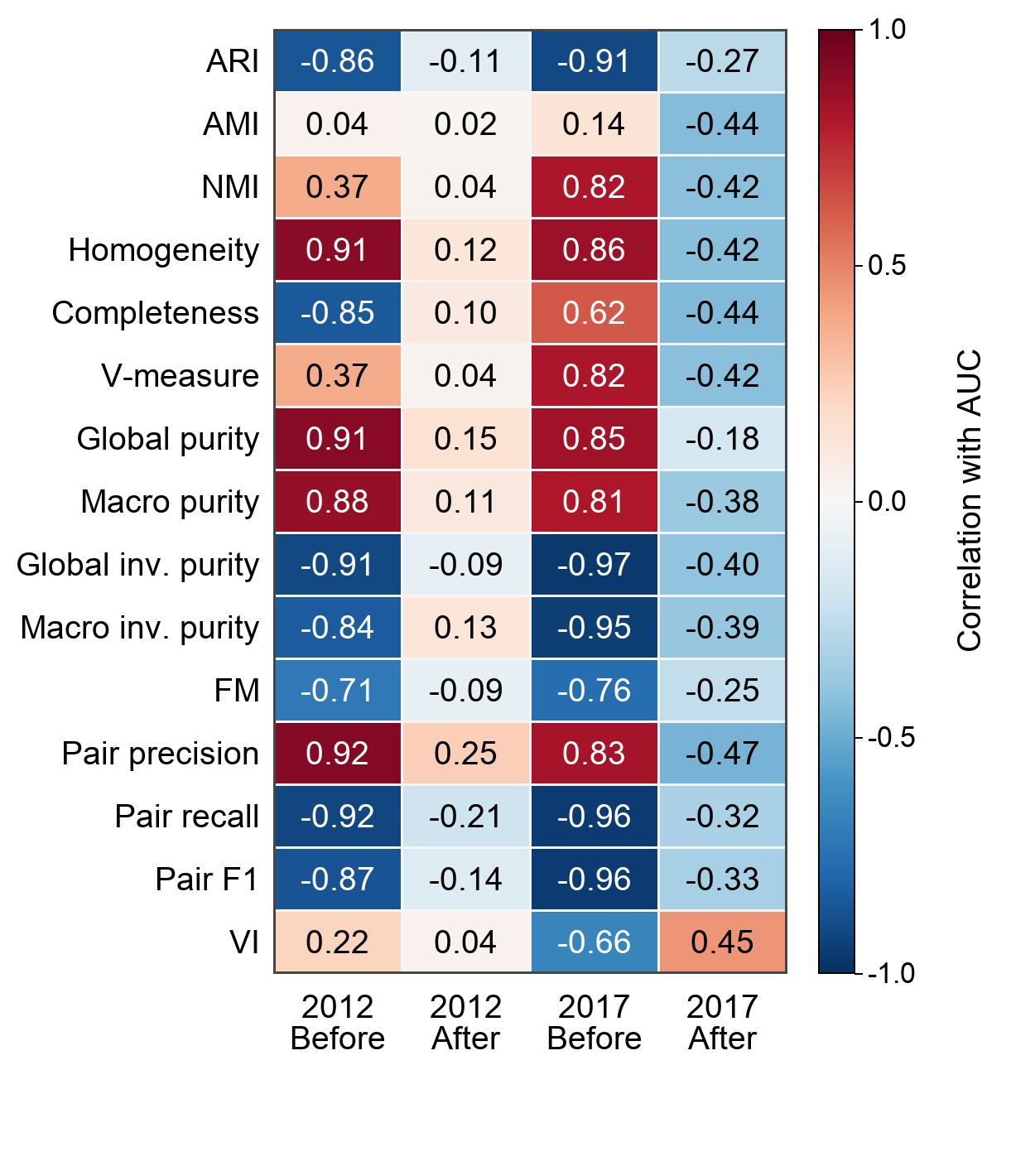}
\caption{Rank correlations between each taxonomy-agreement score and test AUC,
both averaged across the two embedding models. ``After'' indicates that rank
patterns linked to PCA dimension and \(K\) have been removed. Cells show coefficients;
Table~\ref{tab:auc_correlations} reports significance after correction.}
\label{fig:auc_correlations}
\end{figure}

\begin{table}[!t]
\centering
\scriptsize
\setlength{\tabcolsep}{2.2pt}
\resizebox{\columnwidth}{!}{%
\begin{tabular}{lrrrr}
\toprule
Measure & 2012 before & 2012 after & 2017 before & 2017 after \\
\midrule
ARI & \(-.858^{*}\) & \(-.109\) & \(-.908^{*}\) & \(-.273\) \\
AMI & \(.041\) & \(.024\) & \(.138\) & \(-.436^{*}\) \\
NMI & \(.374^{*}\) & \(.038\) & \(.816^{*}\) & \(-.418^{*}\) \\
Homogeneity & \(.906^{*}\) & \(.125\) & \(.857^{*}\) & \(-.425^{*}\) \\
Completeness & \(-.847^{*}\) & \(.097\) & \(.620^{*}\) & \(-.443^{*}\) \\
V-measure & \(.374^{*}\) & \(.038\) & \(.816^{*}\) & \(-.418^{*}\) \\
Global purity & \(.913^{*}\) & \(.150\) & \(.851^{*}\) & \(-.179\) \\
Macro purity & \(.879^{*}\) & \(.105\) & \(.810^{*}\) & \(-.376^{*}\) \\
Global inverse purity & \(-.908^{*}\) & \(-.094\) & \(-.968^{*}\) & \(-.403^{*}\) \\
Macro inverse purity & \(-.836^{*}\) & \(.126\) & \(-.946^{*}\) & \(-.387^{*}\) \\
Fowlkes--Mallows & \(-.714^{*}\) & \(-.088\) & \(-.763^{*}\) & \(-.246\) \\
Pairwise precision & \(.922^{*}\) & \(.246\) & \(.833^{*}\) & \(-.470^{*}\) \\
Pairwise recall & \(-.917^{*}\) & \(-.207\) & \(-.959^{*}\) & \(-.324\) \\
Pairwise \(F_1\) & \(-.866^{*}\) & \(-.139\) & \(-.960^{*}\) & \(-.330\) \\
VI & \(.224\) & \(.040\) & \(-.664^{*}\) & \(.446^{*}\) \\
\bottomrule
\end{tabular}}
\caption{Associations, across the 48 shared settings of PCA dimension and \(K\) in
each dataset, between each taxonomy-agreement score and test AUC, both averaged
across the two embedding models. ``Before'' is the direct Spearman correlation;
``After'' is the correlation after removing rank patterns linked to PCA dimension and \(K\). An
asterisk marks \(p < .05\) after the Benjamini--Hochberg correction, applied
separately by dataset and analysis stage. VI is a distance measure, so a positive
coefficient indicates that a greater difference between the taxonomies is
associated with higher test AUC.}
\label{tab:auc_correlations}
\end{table}

\end{document}